\documentclass[runningheads]{llncs}

\usepackage{graphicx}
\usepackage{amsmath}
\usepackage{booktabs}      
\usepackage{graphicx}     
\usepackage[table]{xcolor} 
\usepackage{array}         
\usepackage{amssymb} 
\usepackage{float}
\usepackage{multirow}
\usepackage{cite}
\usepackage{marvosym}
\usepackage[colorlinks=true, linkcolor=blue, urlcolor=blue]{hyperref}

\begin{document}
	
	\title{FlowDet: Overcoming Perspective and Scale Challenges in Real-Time End-to-End Traffic Detection}
	\titlerunning{FlowDet: A Real-Time End-to-End Traffic Detection}

	\author{
		Zixing Wang \and
		Yuhang Zhao\textsuperscript{(\Letter)}
	}
	
	\institute{
		Chinese Academy of Medical Sciences and Peking Union Medical College, Beijing 100730, China \\
		\email{s2023002057@pumc.edu.cn} \\
		CICT Mobile Communication Technology Co.,Ltd, Beijing 100191, China \\
		\email{ewanzhao@163.com}
	}
	
	\maketitle
	
	\begin{abstract}
		End-to-end object detectors offer a promising NMS-free para-digm for real-time applications, yet their high computational cost remains a significant barrier, particularly for complex scenarios like intersection traffic monitoring. To address this challenge, we propose \textbf{FlowDet}, a high-speed detector featuring a decoupled encoder optimization strategy applied to the DETR architecture. Specifically, FlowDet employs a novel \textbf{Geometric Deformable Unit (GDU)} for traffic-aware geometric modeling and a \textbf{Scale-Aware Attention (SAA)} module to maintain high representational power across extreme scale variations. To rigorously evaluate the model's performance in environments with severe occlusion and high object density, we collected the \textbf{Intersection-Flow-5k} dataset, a new challenging scene for this task. Evaluated on Intersection-Flow-5k, FlowDet establishes a new state-of-the-art. Compared to the strong RT-DETR baseline, it \textbf{improves AP\textsuperscript{test} by 1.5\% and AP\textsubscript{50}\textsuperscript{test} by 1.6\%}, while simultaneously \textbf{reducing GFLOPs by 63.2\% and increasing inference speed by 16.2\%}. Our work demonstrates a new path towards building highly efficient and accurate detectors for demanding, real-world perception systems. The Intersection-Flow-5k dataset is available at 
		\href{https://github.com/AstronZh/Intersection-Flow-5K}{\textcolor{blue}{https://github.com/AstronZh/Intersection-Flow-5K}}.
	\end{abstract}
	
	\section{Introduction}
	
	Real-time traffic flow monitoring at urban intersections is a cornerstone of modern Intelligent Transportation Systems (ITS). Deploying accurate and efficient object detectors on resource-constrained edge devices is paramount for enabling proactive traffic management and automated safety protocols. However, this application domain exposes a fundamental dilemma in current object detection paradigms. On one hand, CNN-based detectors, epitomized by the YOLO series~\cite{YOLO, YOLOv3, YOLOv4, YOLOv7, YOLOv9, YOLOv10}, achieve exceptional inference speed but rely on Non-Maximum Suppression (NMS). This post-processing step becomes an Achilles' heel in crowd-ed intersections, where it frequently fails, erroneously suppressing valid detections of densely packed or occluded vehicles~\cite{soft-nms, YOLOv10}. 
	
	On the other hand, Transformer-based end-to-end detectors like DETR~\cite{EDE} eliminate the NMS bottleneck via set-based prediction. Yet, their prohibitive computational complexity and slow convergence, stemming from global self-attention mechanisms, have historically precluded real-time deployment~\cite{Deformable_DETR}. While recent models like RT-DETR have made significant strides in bridging this efficiency gap~\cite{DETRs}, they remain general-purpose architectures. They are not explicitly engineered to tackle the trifecta of challenges endemic to traffic surveillance: \textbf{1) severe perspective distortion}, \textbf{2) extreme object scale variations}, and \textbf{3) persistent, complex inter-object occlusions}, which are well-documented challenges in modern datasets~\cite{nuScenes, BDD100K}.
	
	To bridge this critical gap, we propose \textbf{FlowDet}, a high-speed end-to-end detection framework specifically engineered for the rigors of traffic intersection monitoring. Our framework is distinguished by two synergistic innovations designed to directly counteract the aforementioned challenges. First, a \textbf{Geometric Deformable Unit (GDU)} that moves beyond standard deformable convolutions~\cite{Deformable_DETR} by adaptively learning traffic-aware sparse sampling points. This allows the model to intrinsically model geometric transformations and focus on discriminative, non-occluded vehicle parts. Second, a \textbf{Scale-Aware Attention (SAA)} mechanism that efficiently processes vast scale disparities through a dual-branch architecture, dedicating parallel pathways to capture both fine-grained local details and broad global context, offering an efficient alternative to classical feature pyramids~\cite{FPN}.
	
\begin{figure}[H]
		\centering
		\includegraphics[width=1\linewidth]{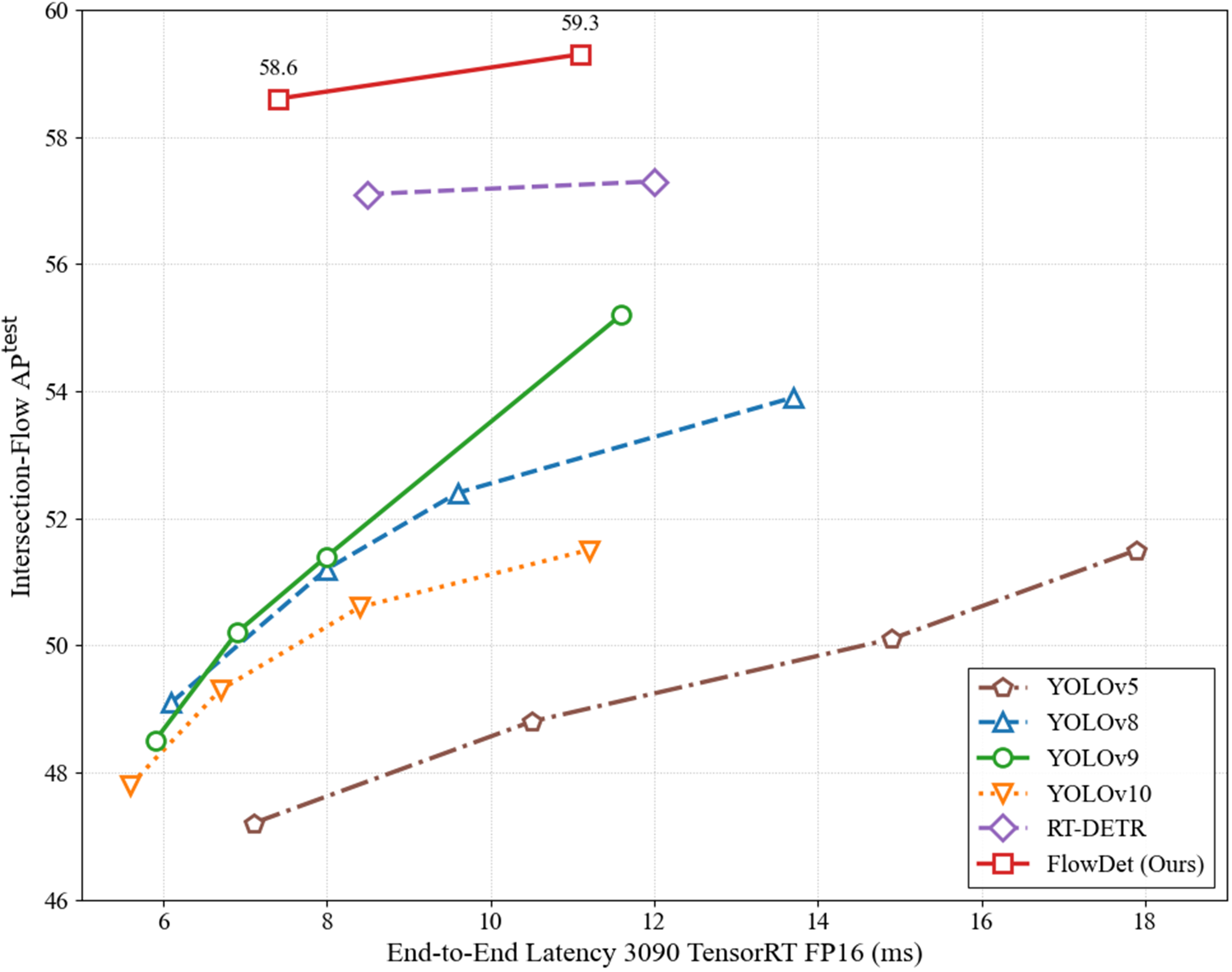}
		\caption{Performance comparison on the Intersection-Flow-5K dataset. FlowDet establishes a new state-of-the-art Pareto curve, achieving superior accuracy at real-time speeds compared to advanced detectors like YOLOv9~\cite{YOLOv9} and RT-DETR~\cite{DETRs}.}
		\label{fig:pareto_curve}
	\end{figure}

	Concurrently, to catalyze research and establish a standardized evaluation platform for this challenging domain, we introduce the \textbf{Intersection-Flow-5K} dataset. This comprehensive scene, comprising 6,928 high-resolution images with over 406,000 meticulously annotated bounding boxes, is the first of its kind specifically focused on complex intersection scenarios. As demonstrated in Figure~\ref{fig:pareto_curve}, extensive experiments show that FlowDet not only establishes a new state-of-the-art on this benchmark and others like COCO~\cite{COCO}, but also maintains the computational efficiency requisite for practical edge deployment. 
	
	Our contributions are summarized as follows:
	\begin{itemize}
		\item \textbf{A novel and efficient end-to-end framework, FlowDet}, specifically designed to address the key challenges of intersection monitoring. It sets a new state-of-the-art in detection accuracy while maintaining real-time, edge-compatible performance.
		\item \textbf{Two synergistic technical innovations}: the \textbf{GDU} for adaptive modeling of perspective and occlusion, and the \textbf{SAA} module for robust detection across extreme scale variations.
		\item \textbf{A new comprehensive scene, Intersection-Flow-5K}, densely annotated with over 400K instances. It serves as a benchmark to facilitate systematic evaluation and foster future research in intersection traffic surveillance.
	\end{itemize}
	
	\section{Related Work}
	\subsection{CNN-based Real-time Detection}
	The YOLO (You Only Look Once) family of detectors has become synonymous with real-time object detection since its initial version introduced a pioneering one-stage CNN architecture~\cite{YOLO}. Through continuous evolution, the YOLO series has consistently outperformed other one-stage approaches, such as SSD~\cite{SSD}, and established dominance in practical applications. This development trajectory has encompassed a wide range of architectural innovations, including both anchor-based and anchor-free paradigms, to achieve strong trade-offs between speed and accuracy. The series continues to advance the state-of-the-art with its most recent iterations~\cite{YOLOv3, YOLOv4, YOLOX, YOLOv7, YOLOv9, YOLOv10}.
	
	The success of YOLO architectures largely stems from efficient feature aggregation mechanisms and backbone designs. Cross Stage Partial (CSP) connections~\cite{CSPNET} have proven particularly effective in YOLO architectures, enabling efficient gradient flow while reducing computational redundancy. CSP networks partition feature maps into two parts, with one part undergoing a series of operations while the other maintains identity connections, ultimately concatenating both parts to create rich feature representations with improved efficiency.
	
	However, these advanced real-time detectors produce numerous overlapping bounding boxes and require Non-Maximum Suppression (NMS) post-processing, which introduces latency bottlenecks in dense scenarios typical of traffic intersections. This fundamental limitation motivates the exploration of end-to-end detection approaches that eliminate post-processing requirements.
	
	\subsection{End-to-End Detection and Efficient Attention}
	Vision Transformers have revolutionized computer vision by using self-attention mechanisms from natural language processing~\cite{transformer}. Since the introduction of ViT for image classification~\cite{imageworth}, subsequent works have enhanced transformer architectures by incorporating convolutional layers, introducing pyramid feature maps, and improving locality through various design innovations~\cite{SwinTF, PVTF}.
	
	DETR~\cite{EDE} introduced groundbreaking end-to-end object detection using Transformers, framing detection as a set prediction problem to eliminate NMS dependency. However, early transformer-based detectors suffered from high computational complexity and slow convergence, limiting practical deployment. Subsequent improvements, such as Deformable DETR~\cite{Deformable_DETR}, focused on improving convergence speed, reducing computational overhead, and improving detection accuracy through architectural innovations~\cite{DAB_DETR, Efficient_DETR}.
	
	Efficient attention mechanisms aim to reduce the quadratic complexity of standard multi-head self-attention (MSA). Representative approaches in computer vision include spatial reduction attention~\cite{PVTF}, local window attention~\cite{SwinTF}, and various hybrid mechanisms. However, existing methods typically focus on either local or global attention patterns within individual layers, leading to suboptimal performance when handling extreme scale variations. Some approaches attempt to address this limitation by introducing additional global tokens or mixing local attention with convolutional operations, but often suffer from inefficient implementations that cannot fully leverage modern hardware acceleration.
	
	RT-DETR~\cite{DETRs} achieved a critical breakthrough by designing an Efficient Hybrid Encoder with optimized query selection mechanisms, successfully bringing end-to-end detection into real-time domains. This advancement established a strong foundation for practical transformer-based detection systems, demonstrating through direct comparisons~\cite{DETRs} that careful architectural design can overcome traditional efficiency limitations of attention-based approaches.
	
	\section{Methodology}
	
	To address the challenges inherent in intersection traffic monitoring, we introduce \textbf{FlowDet}, a detection framework that enhances feature representation through traffic-specific innovations. The framework integrates two core components: a \textbf{Progressive Adaptive Feature Cascade (PAFC)} module featuring our novel \textbf{GDU}, and a \textbf{SAA} module designed for extreme scale variation handling.
	
	\begin{figure}[t]
		\centering
		\includegraphics[width=1\linewidth]{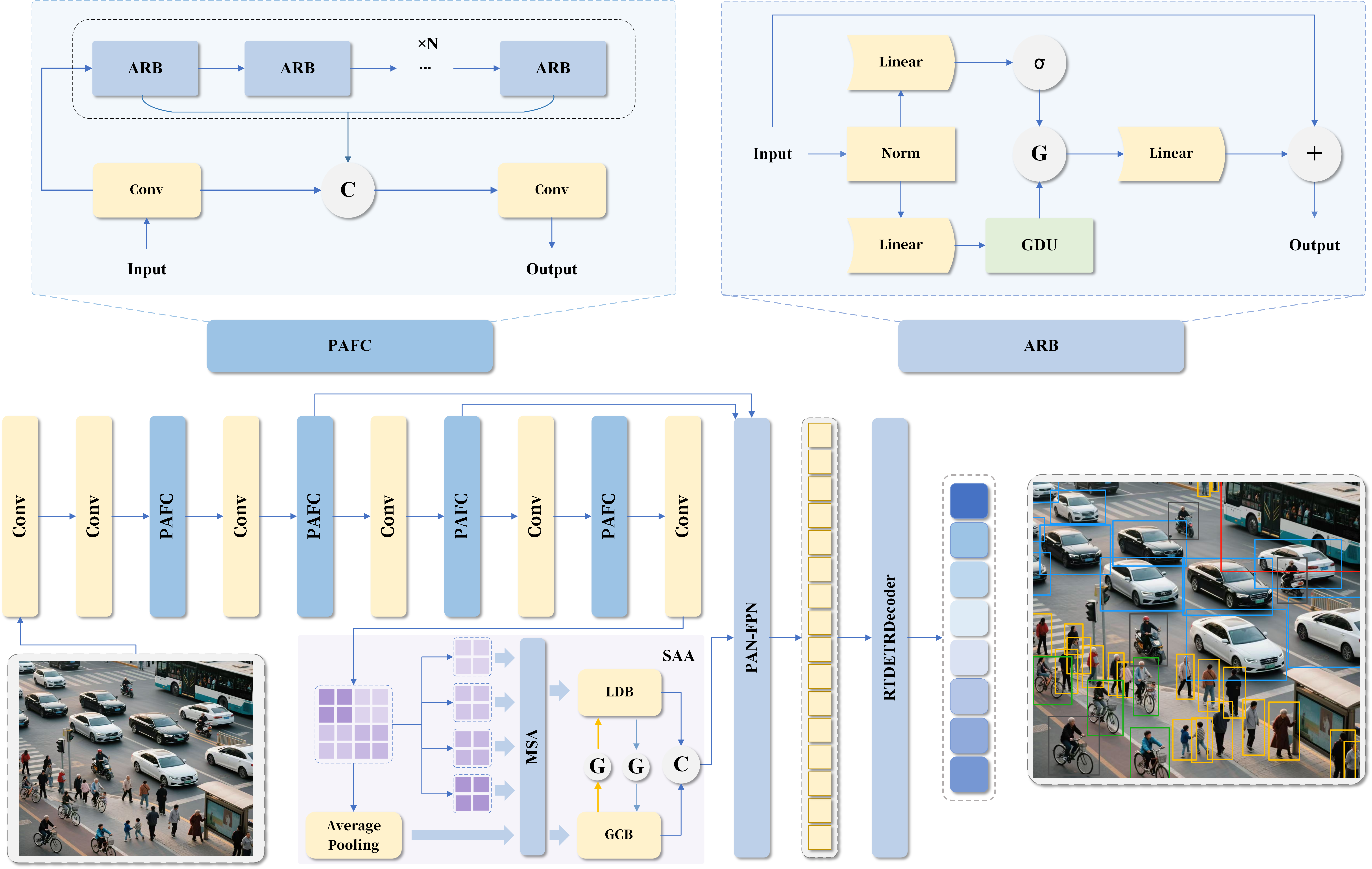}
		\caption{Overall architecture of FlowDet. The framework integrates PAFC with GDU in the backbone and SAA in the encoder for efficient traffic object detection.}
		\label{fig:overall_arch}
	\end{figure}
	
	The synergistic design significantly elevates detection accuracy and robustness while preserving real-time performance essential for practical deployment. Figure~\ref{fig:overall_arch} illustrates the complete architecture.
	
	\subsection {PAFC with Geometric Deformable Unit}
	
	Intersection traffic monitoring presents unique geometric challenges due to \textbf{perspective distortion} caused by fixed camera viewpoints, where vehicles exhibit significant shape variations and spatial transformations across different image regions. Standard backbone architectures are constrained by static convolution operations, limiting their ability to model these geometric variations effectively. The PAFC module addresses this through specialized \textbf{Adaptive Refinement Blocks (ARBs)} that progressively refine features while dynamically adjusting receptive fields to accommodate perspective-induced deformations.
	
	The PAFC architecture employs cross-stage partial connections for efficient information flow:
	
	\begin{equation}\label{eq:pafc_init}
		\mathbf{Y}_0, \mathbf{Y}_1 = \text{Split}\left(Conv(\mathbf{X})\right)
	\end{equation}
	
	\begin{equation}\label{eq:pafc_refinement}
		\mathbf{Y}_{i} = \text{ARB}_{\text{i}}(\mathbf{Y}_{i-1}) , \quad \text{for } i \in \{2, \ldots, n\}
	\end{equation}
	
	\begin{equation}\label{eq:pafc_aggregation}
		\mathbf{Y}_{\text{PAFC}} = \mathcal{F}_{\text{fusion}}\left( \bigoplus_{i=0}^{n} \mathcal{W}_i \mathbf{Y}_i \right)
	\end{equation}
	
	Unlike standard deformable convolutions that learn unconstrained 2D offsets, the GDU disentangles the modeling of complex geometric transformations. It features two parallel deformable convolution branches, termed Horizontality and Verticality, which learn spatial offsets primarily along their respective principal axes. This specialized design more effectively captures the systematic shearing and foreshortening effects prevalent in traffic surveillance imagery.
	
	A critical component of the GDU is the modulation term $\psi(\|\Delta\mathbf{p}_k^{\text{geo}}\|_2)$, as shown in Eq.~\eqref{eq:gdu_operation}. This term dynamically re-weights the contribution of each sampled point based on the magnitude of its learned offset ($\Delta\mathbf{p}_k^{\text{geo}}$). This allows the model to prioritize stable, nearby points (small offsets) while cautiously sampling from distant, potentially noisy locations (large offsets), significantly improving the robustness of the feature representation.
	
	The GDU operation is formulated as:
	
	\begin{equation}\label{eq:gdu_operation}
		\text{GDU}(\mathbf{X}; \mathbf{p}_0) = \sum_{k \in \mathcal{K}} \omega_k^{\text{geo}} \cdot \mathbf{X}(\mathbf{p}_0 + \mathbf{p}_k + \Delta\mathbf{p}_k^{\text{geo}}) \cdot \psi(\|\Delta\mathbf{p}_k^{\text{geo}}\|_2)
	\end{equation}
	
	The geometric offset prediction incorporates both local and global context to \textbf{address perspective distortion systematically}:
	
	\begin{equation}\label{eq:gdc_offset}
		\{\Delta\mathbf{p}_k\}_{k \in \mathcal{K}} = \mathcal{F}_{\text{offset}}(\text{DWConv}(\mathbf{X})) \cdot \sigma
	\end{equation}
	
	Here, $\text{DWConv}$ (Depth-wise Separable Convolution\cite{DWconv}) is used to efficiently capture spatial patterns for offset prediction without significant computational overhead. The scaling factor $\sigma$ modulates the magnitude of the offsets, ensuring training stability.
	
	\begin{figure}[t]
		\centering
		\includegraphics[width=0.7\linewidth]{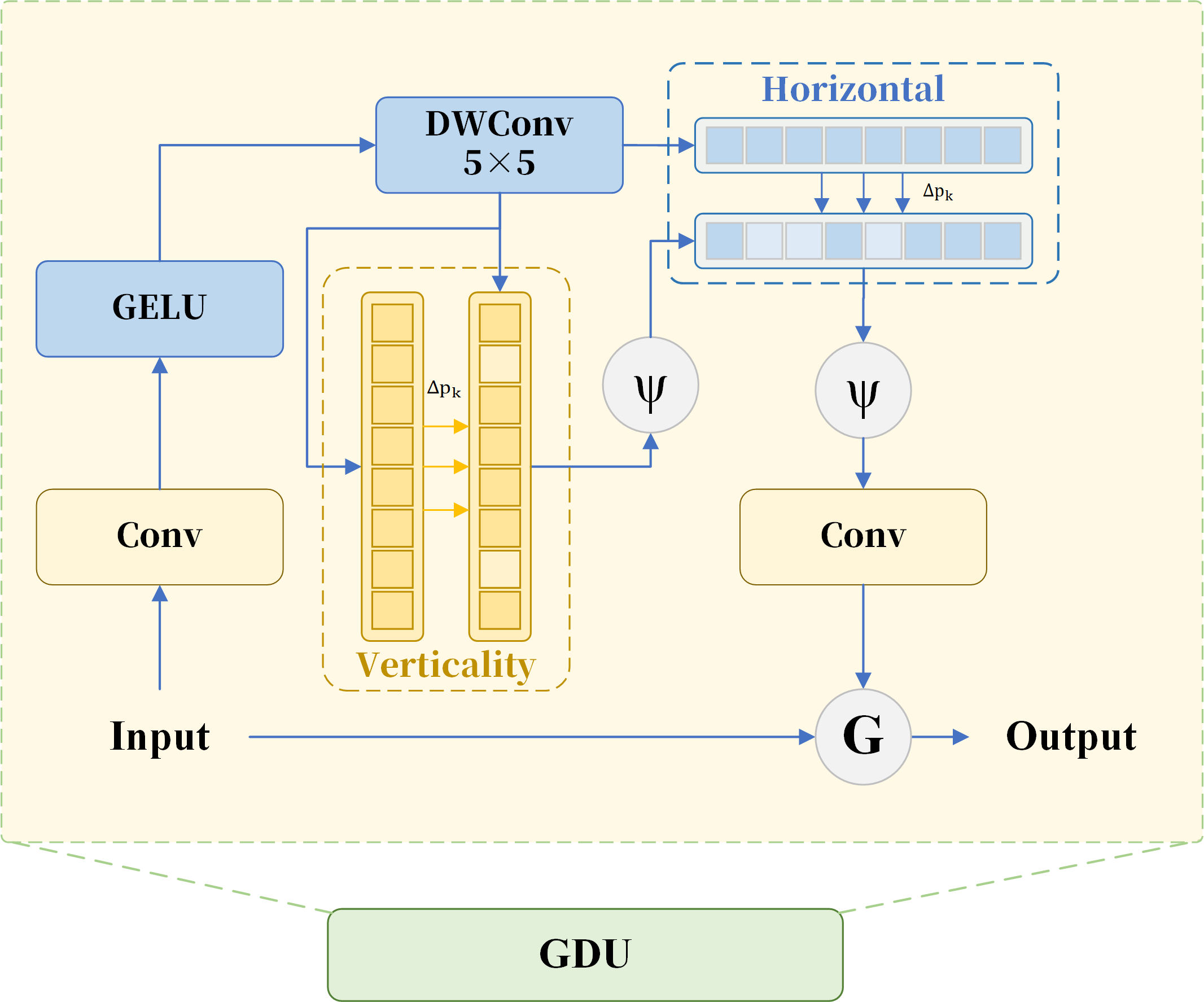}
		\caption{GDU within Adaptive Refinement Unit. The learned geometric-aware spatial offsets enable adaptive modeling for traffic objects with varying perspective characteristics.}
		\label{fig:gdu}
	\end{figure}
	
	\subsection{Scale-Aware Attention (SAA)}
	
	Traffic monitoring demands simultaneous processing across extreme scale variations, from \textbf{distant vehicles occupying minimal pixel areas to nearby large objects}. Traditional attention mechanisms face fundamental trade-offs between capturing fine-grained local texture details essential for small object detection and modeling global contextual relationships necessary for scene understanding. The SAA module employs a \textbf{dual-branch architecture with adaptive gate-controlled fusion} to decouple these processes while maintaining computational efficiency.
	
	The SAA processes input features through specialized branches with \textbf{learnable gate mechanisms} that adaptively balance local and global information:
	
	\begin{equation}\label{eq:saa_architecture}
		\mathbf{F}_{\text{out}} = \mathcal{F}_{\text{fusion}}\left(\text{LDB}(\mathbf{F}_{\text{in}}), \text{GCB}(\mathbf{F}_{\text{in}}), \mathcal{W}_{\text{gate}}(\mathbf{F}_{\text{in}})\right)
	\end{equation}
	
	\subsubsection{Local Detail Branch (LDB) for Small Object Optimization}
	
	Our Local Detail Block (LDB) first partitions the input feature map into non-overlapping windows. 
	Within each window, we apply self-attention enhanced by our specific Local Position Encoding to effectively model local geometric relationships.
	\begin{equation}\label{eq:ldb_attention}
		\text{WindowAttn}(\mathbf{W}) = \text{Softmax}\left(\frac{\mathbf{Q}_{\text{window}}\mathbf{K}_{\text{window}}^T}{\sqrt{d_k}} + \text{LPE}_{\text{specific}}\right) \mathbf{V}_{\text{window}}
	\end{equation}
	The processed windows are then merged back to form the final output feature map. This windowed attention mechanism ensures computational efficiency while preserving the high spatial resolution crucial for localizing small, distant vehicles. By focusing on a constrained local context, the LDB excels at discriminating fine-grained features, such as vehicle contours and lights, which are vital for small object detection.
	
	\subsubsection{Global Context Branch (GCB) for Scene Understanding}
	
	The Global Context Branch models long-range dependencies and \textbf{scene-level contextual relationships} through spatial reduction attention:
	
	\begin{equation}\label{eq:gcb_attention}
		\text{GlobalAttn}(\mathbf{F}) = \text{Softmax}\left(\frac{\mathbf{Q}_{\text{global}}\mathbf{K}_{\text{global}}^T}{\sqrt{d_k}} + \text{GPE}_{\text{scene}}(\mathbf{F})\right) \mathbf{V}_{\text{global}}
	\end{equation}
	
	GCB employs global position encodings ($\text{GPE}_{\text{scene}}$) that capture \textbf{inter-object spatial relationships and scene structure}, enabling robust detection in cluttered environments with complex occlusion patterns.
	
	\subsubsection{Adaptive Gate-Controlled Fusion}
	
	The \textbf{gate mechanism dynamically balances local texture details and global contextual information} based on spatial characteristics and object scales:
	
	\begin{equation}\label{eq:saa_fusion}
		\mathcal{F}_{\text{fusion}} = \mathbf{F}_{\text{local}} \odot (1 - \mathcal{W}_{\text{gate}}) + \mathbf{F}_{\text{global}} \odot \mathcal{W}_{\text{gate}} + \mathbf{F}_{\text{cross}}
	\end{equation}
	
	where $\mathcal{W}_{\text{gate}}$ learns to \textbf{emphasize local details for small objects while prioritizing global context for large objects and scene understanding}. The cross-scale interaction term $\mathbf{F}_{\text{cross}}$ facilitates information exchange between branches, ensuring coherent multi-scale feature representation.
	
	This adaptive fusion mechanism enables SAA to \textbf{automatically adjust attention focus}: emphasizing fine-grained local texture patterns when processing small/distant objects, while leveraging global contextual relationships for understanding complex traffic scenarios and resolving occlusions. The gate weights are learned end-to-end, allowing the model to develop traffic-specific attention strategies optimized for intersection monitoring challenges.
	
	\section{Intersection-Flow-5K Dataset}
	
	Traffic intersection monitoring presents unique challenges inadequately addressed by existing benchmarks. Infrastructure-based cameras observe persistent occlusion patterns, extreme scale variations, and complex illumination conditions that differ fundamentally from ego-centric driving datasets. To address these limitations, we introduce \textbf{Intersection-Flow-5K}, a specialized benchmark designed for real-world traffic surveillance evaluation.
	
	\begin{table}[H]
		\centering
		\caption{Intersection-Flow-5K Dataset Statistics}
		\label{tab:dataset_stats}
		\scriptsize
		\begin{tabular}{l|c|c|c|c}
			\toprule
			\textbf{Category} & \textbf{Train} & \textbf{Val} & \textbf{Test} & \textbf{Total} \\
			\midrule
			Images & 5,483 & 722 & 723 & 6,928 \\
			\midrule
			Vehicle & 190,417 & 23,450 & 23,978 & 237,845 \\
			Bus & 2,816 & 315 & 322 & 3,453 \\
			Bicycle & 23,396 & 2,771 & 3,051 & 29,218 \\
			Pedestrian & 19,005 & 2,250 & 2,459 & 23,714 \\
			Engine & 642 & 74 & 80 & 796 \\
			Truck & 7,749 & 881 & 916 & 9,546 \\
			Tricycle & 1,610 & 225 & 200 & 2,035 \\
			Obstacle & 82,190 & 8,902 & 9,059 & 100,151 \\
			\midrule
			\textbf{Total Objects} & \textbf{327,825} & \textbf{38,868} & \textbf{40,065} & \textbf{406,758} \\
			\bottomrule
		\end{tabular}
	\end{table}
	
	The dataset construction involved systematic sampling from urban intersections, comprising 6,928 high-resolution images (1920×1080) with over 406,000 annotations across 8 categories. Table~\ref{tab:dataset_stats} presents comprehensive statistics demonstrating balanced representation across training, validation, and test splits. The annotation protocol mandates labeling of heavily occluded objects (up to 75\% occlusion) and extremely small objects (minimum 15×15 pixels), essential for realistic traffic monitoring evaluation.
	
\begin{figure}[H]
		\centering
		\includegraphics[width=0.8\linewidth]{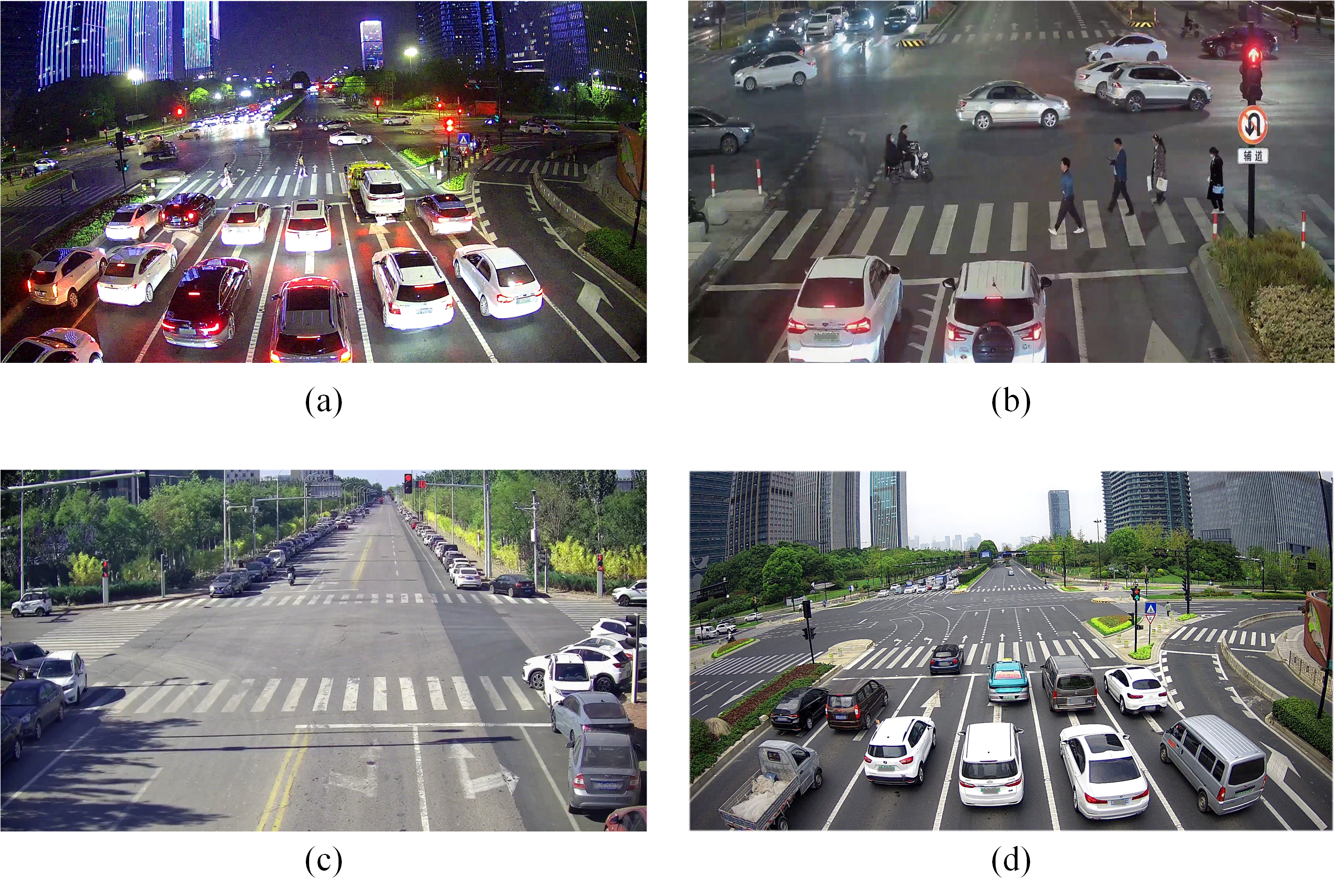}
		\caption{Intersection-Flow-5K captures a spectrum of critical challenges in traffic surveillance, including (a) sensor saturation from nighttime glare, (b) complex inter-object occlusion, (c) extreme scale variation with distant objects, and (d) baseline ideal conditions.}
		\label{fig:dataset_examples}
	\end{figure}

	Figure~\ref{fig:dataset_examples} illustrates four critical scenarios that characterize intersection monitoring challenges. The nighttime glare scenario demonstrates direct headlight interference that saturates sensors and obscures nearby objects. The vehicle occlusion scenario shows persistent blockage where large vehicles systematically obscure smaller traffic participants in adjacent lanes. The small object scenario captures distant vehicles at minimal pixel resolutions, testing detection capabilities at extreme scales. The ideal scenario provides baseline comparison under optimal visibility conditions. These scenarios establish comprehensive evaluation protocols for traffic-specific detection algorithms.
	\section{Experiments}
	
	All experiments are conducted on RTX 3090 GPUs under TensorRT FP16 precision with batch size 1. We adopt unified 640×640 input resolution for fair comparison across all methods. Training employs AdamW optimizer with learning rate 1e-4, weight decay 1e-4, and cosine scheduling over 200 epochs.
	
	\subsection{Comparison with State-of-the-Art Methods}
	
	We compare FlowDet against representative methods from real-time and end-to-end detection paradigms, organizing results into complementary tables for clarity.
	
	\begin{table}[t]
		\centering
		\caption{Comparison with Real-Time Detection Models. All methods use unified 640×640 input resolution for fair comparison under identical experimental conditions.}
		\label{tab:realtime_comparison}
		\scriptsize
		\setlength{\tabcolsep}{3pt}
		\begin{tabular}{l|c|c|c|c|c|c|c|c}
			\toprule
			\textbf{Method} & \textbf{Epochs} & \textbf{Params} & \textbf{GFLOPs} & \textbf{FPS} & \textbf{AP$^{test}$} & \textbf{AP$_{50}^{test}$ } & \textbf{AP$_{S}^{test}$} & \textbf{AP$_{L}^{test}$} \\
			& & \textbf{(M)} & & & & & & \\
			\midrule
			YOLOv5-L  & 400 & 46 & 109 & 67 & 50.1 & 68.8 & 19.1 & 72.3 \\
			YOLOv5-X  & 400 & 86 & 205 & 56 & 51.5 & 70.2 & 20.3 & 73.8 \\
			YOLOv8-L  & 400 & 43 & 165 & 104 & 52.4 & 71.7 & 20.0 & 87.4 \\
			YOLOv8-X  & 400 & 68 & 257 & 73 & 53.9 & 73.1 & 29.5 & 74.3 \\
			YOLOv9-C  & 400 & 25 & 102 & 125 & 51.4 & 70.5 & 18.9 & 86.9 \\
			YOLOv9-E  & 400 & 57 & 189 & 86 & 55.2 & 73.8 & 25.5 & 88.0 \\
			YOLOv10-L & 400 & 24 & 120 & 119 & 50.6 & 69.8 & 19.5 & 74.2 \\
			YOLOv10-X & 400 & 29 & 160 & 89 & 51.5 & 70.5 & 18.4 & 74.9 \\
			Faster-RCNN & 600 & 42 & 180 & 18 & 45.2 & 65.1 & 25.4 & 60.5 \\
			Faster-RCNN & 600 & 60 & 247 & 14 & 46.8 & 67.0 & 27.0 & 62.1 \\
			\midrule
			\textbf{FlowDet}  & \textbf{200} & \textbf{15} & \textbf{50} & \textbf{136} & \textbf{58.6} & \textbf{82.3} & \textbf{34.2} & \textbf{89.0} \\
			\textbf{FlowDet-L} & \textbf{200} & \textbf{21} & \textbf{78} & \textbf{90} & \textbf{59.3} & \textbf{83.1} & \textbf{35.4} & \textbf{89.7} \\
			\bottomrule
		\end{tabular}
	\end{table}
	
	\begin{table}[t]
		\centering
		\caption{Comparison with End-to-End Detection Models. All transformer-based methods evaluated under identical experimental protocols and hardware configurations.}
		\label{tab:endtoend_comparison}
		\scriptsize
		\setlength{\tabcolsep}{3pt}
		\begin{tabular}{l|c|c|c|c|c|c|c|c}
			\toprule
			\textbf{Method} & \textbf{Epochs} & \textbf{Params} & \textbf{GFLOPs} & \textbf{FPS} & \textbf{AP$^{test}$} & \textbf{AP$_{50}^{test}$} & \textbf{AP$_{S}^{test}$} & \textbf{AP$_{L}^{test}$} \\
			& & \textbf{(M)} & & & & & & \\
			\midrule
			DETR-DC5 & 500 & 41 & 187 & 10 & 47.5 & 67.3 & 25.8 & 66.2 \\
			DETR-DC5 & 500 & 60 & 253 & 7 & 49.1 & 68.9 & 27.4 & 67.8 \\
			Deformable-DETR & 300 & 40 & 173 & 15 & 52.8 & 72.5 & 33.1 & 68.3 \\
			DAB-Deformable-DETR & 300 & 48 & 195 & 12 & 53.9 & 73.4 & 34.0 & 69.5 \\
			Efficient-DETR & 150 & 35 & 210 & 25 & 50.8 & 70.5 & 30.1 & 67.9 \\
			Efficient-DETR & 150 & 54 & 289 & 17 & 52.1 & 71.8 & 31.5 & 69.3 \\
			RT-DETR-R50 & 200 & 42 & 136 & 117 & 57.1 & 80.7 & 31.0 & 85.0 \\
			RT-DETR-R101 & 200 & 76 & 259 & 83 & 57.3 & 81.4 & 32.1 & 87.2 \\
			\midrule
			\textbf{FlowDet} & \textbf{200} & \textbf{15} & \textbf{50} & \textbf{136} & \textbf{58.6} & \textbf{82.3} & \textbf{34.2} & \textbf{89.0} \\
			\textbf{FlowDet-L} & \textbf{200} & \textbf{21} & \textbf{78} & \textbf{90} & \textbf{59.3} & \textbf{83.1} & \textbf{35.4} & \textbf{89.7} \\
			\bottomrule
		\end{tabular}
	\end{table}
	
	FlowDet demonstrates superior performance across all metrics. Compared to RT-DETR-R50, our method achieves \textbf{1.5\% AP$^{test}$ improvement} (58.6\% vs 57.1\%) and \textbf{1.6\% AP$_{50}^{test}$ enhancement} (82.3\% vs 80.7\%) while reducing computational cost by \textbf{63.2\%} (50.0 vs 136.0 GFLOPs) and increasing inference speed by \textbf{16.2\%} (136 vs 117 FPS). Performance improvements are particularly pronounced for small objects (AP$_S^{test}$: 34.2\% vs 31.0\%), critical for traffic monitoring where distant vehicles must be detected reliably.
	
	\subsection{Comprehensive Ablation Studies}
	
	We conduct extensive ablation studies to validate component contributions and analyze design choices.
	
	The primary ablation demonstrates that both SAA and PAFC contribute significantly to performance improvement. SAA alone achieves 1.2\% AP$^{test}$ improvement while reducing latency by 0.5ms. PAFC provides 1.1\% AP$^{test}$ enhancement and 0.7ms latency reduction. Combined, they yield 1.5\% AP$^{test}$ improvement and 1.1ms latency reduction.
	\begin{table}[H]
		\centering
		\caption{Component-wise Ablation Analysis}
		\label{tab:ablation_main}
		\begin{tabular}{c|c|c|c|c|c}
			\toprule
			\textbf{ID} & \textbf{Backbone} & \textbf{Attention} & \textbf{Latency (ms)} & \textbf{AP$^{test}$ (\%)} & \textbf{AP$_{50}^{test}$ (\%)} \\ 
			\midrule
			1 & ResNet-50 & AIFI & 8.5 & 57.1 & 80.7  \\ 
			2 & ResNet-50 & \textbf{SAA} & 8.0 & 58.3 & 81.9  \\ 
			3 & \textbf{PAFCNet} & AIFI & 7.8 & 58.2 & 82.1  \\ 
			4 & \textbf{PAFCNet} & \textbf{SAA} & \textbf{7.4} & \textbf{58.6} & \textbf{82.3}  \\ 
			\bottomrule
		\end{tabular}
	\end{table}

	\begin{table}[H]
		\caption{SAA Branch Weight Sensitivity Analysis}
		\label{tab:saa_analysis}
		\begin{tabular}{c|c|c|c|c|c}
			\toprule
			\textbf{Local Weight} & \textbf{Global Weight} & \textbf{AP$^{test}$ (\%)} & \textbf{AP$_{50}^{test}$ (\%)} & \textbf{AP$_S^{test}$ (\%)} & \textbf{AP$_L^{test}$ (\%)} \\ 
			\midrule
			0.3 & 0.7 & 57.8 & 81.2 & 32.1 & 88.4  \\ 
			0.4 & 0.6 & 58.1 & 81.6 & 33.2 & 88.7  \\ 
			0.5 & 0.5 & 58.6 & 82.3 & 34.2 & 89.0  \\ 
			0.6 & 0.4 & 58.3 & 82.0 & 33.8 & 88.9  \\ 
			0.7 & 0.3 & 57.9 & 81.4 & 33.1 & 88.6  \\ 
			\bottomrule
		\end{tabular}
	\end{table}
	
	Weight sensitivity analysis reveals that balanced local/global processing achi-eves optimal performance, particularly for small object detection (AP$_S^{test}$: 34.2\%). This validates our adaptive fusion mechanism that dynamically adjusts branch contributions.
	
	\begin{table}[H]
		\centering
		\caption{Local Detail Attention Window Size Analysis}
		\label{tab:window_size_analysis}
		\begin{tabular}{c|c|c|c|c|c}
			\toprule
			\textbf{Window Size} & \textbf{Latency (ms)} & \textbf{AP$_S$ (\%)} & \textbf{AP$_M$ (\%)} & \textbf{AP$_L$ (\%)} & \textbf{Overall AP (\%)} \\
			\midrule
			1$\times$1 & 6.8  & 32.1 & 59.2 & 88.7 & 58.1 \\
			2$\times$2 & 7.4  & 34.2 & 59.5 & 89.0 & 58.6 \\
			4$\times$4 & 8.9  & 33.8 & 58.9 & 88.4 & 58.3 \\
			8$\times$8 & 11.2 & 32.5 & 57.8 & 87.9 & 57.7 \\
			\bottomrule
		\end{tabular}
	\end{table}
	
	The window size analysis demonstrates that 2$\times$2 windowing provides an optimal trade-off between local detail capture and computational efficiency. Smaller windows (1$\times$1) lack sufficient context for effective local feature learning, while larger windows (4$\times$4, 8$\times$8) introduce excessive computational overhead without proportional accuracy gains.
	
	\subsection{Qualitative Analysis and Visualization}
	
\begin{figure}[H]
		\centering
		\includegraphics[width=0.85\linewidth]{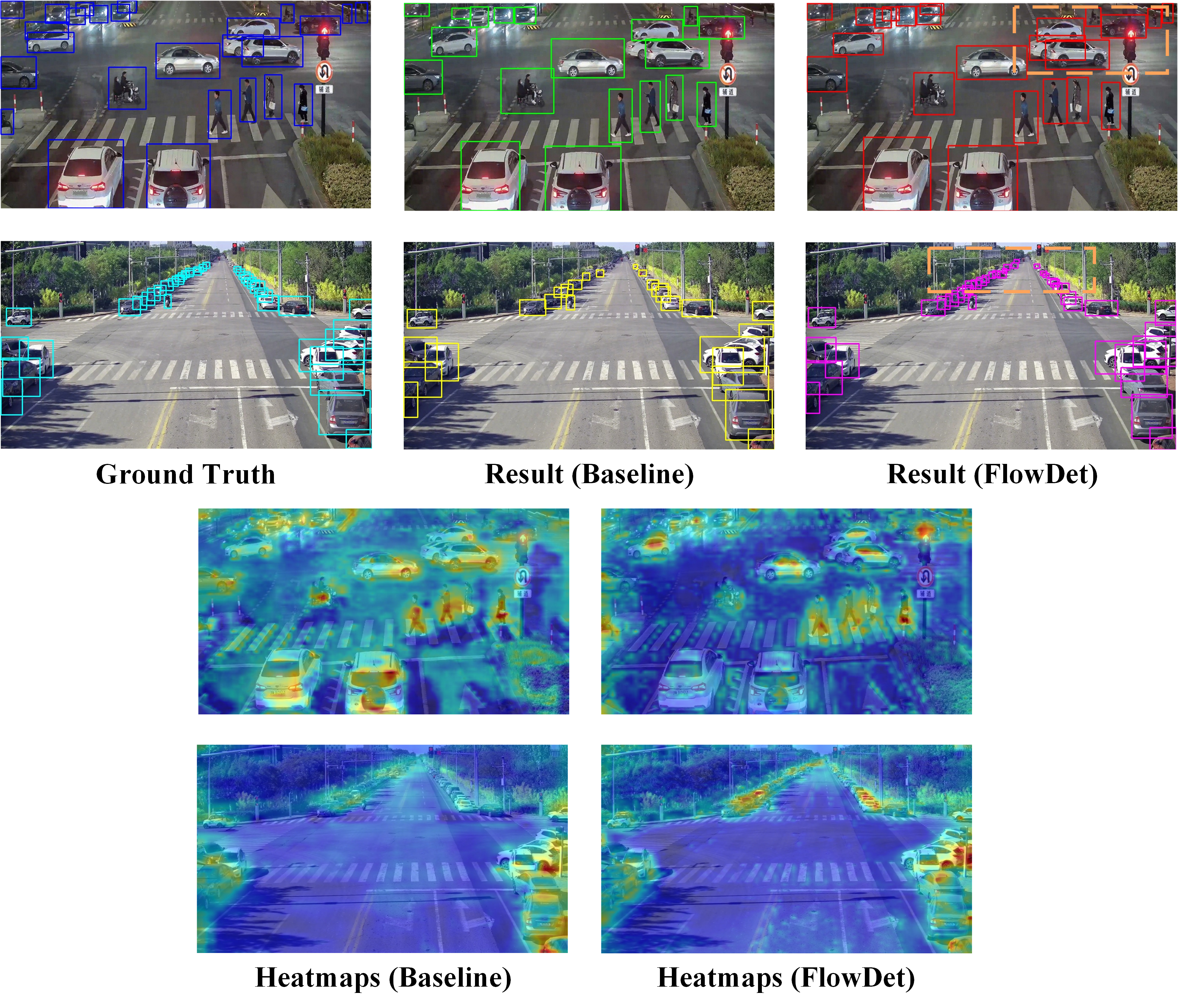}
		\caption{Qualitative comparison showcasing FlowDet's superior attention mechanism and detection performance.
			\textbf{Top row (Occlusion):} In challenging scenarios with severe inter-object occlusion, FlowDet precisely focuses on visible vehicle parts, leading to robust detection. 
			\textbf{Bottom row (Small Objects):} For distant, low-resolution targets, our model maintains a tight attention focus, demonstrating its resilience to extreme scale variations. 
			FlowDet's refined geometric and scale-aware modeling results in more accurate and reliable detections in both cases.}
		\label{fig:qualitative}
	\end{figure}

	Figure~\ref{fig:qualitative} provides qualitative analysis of FlowDet's performance in challenging scenarios. The vehicle occlusion scenarios demonstrate FlowDet's ability to detect partially occluded objects through improved context understanding and geometric modeling capabilities. The attention heatmaps reveal that our SAA mechanism effectively focuses on relevant object regions while suppressing background noise. In small object scenarios, FlowDet consistently detects distant vehicles that baseline methods miss, validating the effectiveness of our scale-aware attention design. The improved attention patterns demonstrate that our dual-branch architecture successfully captures both local details necessary for small object detection and global context required for understanding scene structure and object relationships.
	\section{Conclusion}
	
	This work presents FlowDet, a novel approach for real-time traffic flow detection that enhances DETR architecture through specialized innovations for intersection monitoring. Our framework integrates two key components: the Geometric Deformable Unit enabling adaptive geometric modeling for traffic objects with varying characteristics, and the Scale-Aware Attention module providing efficient multi-scale processing through dual-branch architecture. Comprehensive evaluation on the newly introduced Intersection-Flow-5K dataset demonstrates significant improvements: 1.5\% AP$^{test}$ improvement and 63.2\% computational reduction compared to RT-DETR baseline while maintaining 136 FPS real-time performance. The method shows particular strength in small object detection (34.2\% AP$_S^{test}$ vs 31.0\% baseline), crucial for traffic monitoring applications. Our traffic-specific optimizations make FlowDet suitable for deployment in resource-constrained edge computing scenarios typical of traffic infrastructure, providing practical value for intelligent transportation systems.


\begin{thebibliography}{99}
		\bibitem{BDD100K}
		Yu, F., Chen, H., Wang, X., Xian, W., Chen, Y., Liu, F., Madhavan, V., Darrell, T.: BDD100K: A diverse driving dataset and challenges for open-vocabulary object detection and tracking. In: Proceedings of the IEEE/CVF Conference on Computer Vision and Pattern Recognition. pp. 2636–2645 (2020)
		
		\bibitem{soft-nms}
		Bodla, N., Singh, B., Chellappa, R., Davis, L.S.: Soft-NMS--improving object detection with one line of code. In: Proceedings of the IEEE international conference on computer vision. pp. 5561--5569 (2017)
		
		\bibitem{nuScenes}
		Caesar, H., Bankiti, V., Lang, A.H., Vora, S., Liong, V.E., Xu, Q., Krishnan, A., Pan, Y., Baldan, G., Beijbom, O.: nuScenes: A multimodal dataset for autonomous driving. In: Proceedings of the IEEE/CVF conference on computer vision and pattern recognition. pp. 11621--11631 (2020)
		
		\bibitem{FPN}
		Lin, T.Y., Dollár, P., Girshick, R., He, K., Hariharan, B., Belongie, S.: Feature pyramid networks for object detection. In: Proceedings of the IEEE conference on computer vision and pattern recognition. pp. 2117--2125 (2017)
		
		\bibitem{YOLO}
		Redmon, J., Divvala, S., Girshick, R., Farhadi, A.: You only look once: Unified, real-time object detection. In: Proceedings of the IEEE conference on computer vision and pattern recognition. pp. 779–788 (2016)
		
		\bibitem{SSD} 
		Liu, W., Anguelov, D., Erhan, D., Szegedy, C., Reed, S., Fu, C. Y., Berg, A. C. (2016). SSD: Single shot multibox detector. In: European conference on computer vision. pp. 21–37. Springer (2016)
		
		\bibitem{YOLOv3}
		Redmon, J., Farhadi, A.: YOLOv3: An incremental improvement. arXiv preprint arXiv:1804.02767(2018)
		
		\bibitem{YOLOv4}
		Bochkovskiy, A., Wang, C.Y., Liao, H.Y.M.: YOLOv4: Optimal speed and accuracy of object detection. arXiv preprint arXiv:2004.10934(2020)
		
		\bibitem{YOLOX} 
		Ge, Z., Liu, S., Wang, F., Li, Z., Sun, J. Yolox: Exceeding yolo series in 2021. arXiv preprint arXiv:2107.08430(2021)
		
		\bibitem{YOLOv7}
		Wang, C.Y., Bochkovskiy, A., Liao, H.Y.M.: YOLOv7: Trainable bag-of-freebies sets new state-of-the-art for real-time object detection. In: Proceedings of the IEEE/CVF Conference on Computer Vision and Pattern Recognition. pp. 7464–7475 (2022)
		
		\bibitem{YOLOv9}
		Wang, C.Y., Yeh, I.H., Liao, H.Y.M.: YOLOv9: Learning what you want to learn using programmable gradient information. arXiv preprint arXiv:2402.13616(2024)
		
		\bibitem{YOLOv10}
		Wang, A., Chen, H., Liu, L., Chen, K., Lin, Z., Han, J., Ding, G.: YOLOv10: Real-time end-to-end object detection. arXiv preprint arXiv:2405.14458 (2024)
		
		\bibitem{CSPNET}
		Wang, C.Y., Liao, H.Y.M., Wu, Y.H., Chen, P.Y., Hsieh, J.W., Yeh, I.H.: CSPNet: A new backbone that can enhance learning capability of CNN. In: Proceedings of the IEEE/CVF conference on computer vision and pattern recognition workshops. pp. 390–391 (2020)
		
		\bibitem{transformer}
		Vaswani, A., Shazeer, N., Parmar, N., Uszkoreit, J., Jones, L., Gomez, A.N., Kaiser, Ł., Polosukhin, I.: Attention is all you need. In: Advances in neural information processing systems. pp. 5998–6008 (2017)
		
		\bibitem{imageworth}
		Dosovitskiy, A., Beyer, L., Kolesnikov, A., Weissenborn, D., Zhai, X., Unterthiner, T., Dehghani, M., Minderer, M., Heigold, G., Gelly, S., et al.: An image is worth 16x16 words: Transformers for image recognition at scale. arXiv preprint arXiv:2010.11929 (2020)
		
		\bibitem{SwinTF}
		Liu, Z., Lin, Y., Cao, Y., Hu, H., Wei, Y., Zhang, Z., Lin, S., Guo, B.: Swin transformer: Hierarchical vision transformer using shifted windows. In: Proceedings of the IEEE/CVF International Conference on Computer Vision. pp. 10012–10022 (2021)
		
		\bibitem{PVTF}
		Wang, W., Xie, E., Li, X., Fan, D.P., Song, K., Liang, D., Lu, T., Luo, P., Shao, L.: Pyramid vision transformer: A versatile backbone for dense prediction without convolutions. In: Proceedings of the IEEE/CVF International Conference on Computer Vision. pp. 568–578 (2021)
		
		\bibitem{EDE}
		Carion, N., Massa, F., Synnaeve, G., Usunier, N., Kirillov, A., Zagoruyko, S.: End-to-end object detection with transformers. In: European conference on computer vision. pp. 213–229. Springer (2020)
		
		\bibitem{Deformable_DETR}
		Zhu, X., Su, W., Lu, L., Li, B., Wang, X., Dai, J.: Deformable DETR: Deformable transformers for end-to-end object detection. arXiv preprint arXiv:2010.04159 (2020)
		
		\bibitem{DAB_DETR}
		Liu, S., Li, F., Zhang, H., Yang, X., Qi, X., Su, H., Zhu, J., Zhang, L.: DAB-DETR: Dynamic anchor boxes are better queries for DETR. arXiv preprint arXiv:2201.12329 (2022)
		
		\bibitem{Efficient_DETR}
		Yao, Z., Ai, J., Li, B., Zhang, C.: Efficient DETR: improving end-to-end object detector with dense prior. arXiv preprint arXiv:2104.01318 (2021)
		
		\bibitem{DETRs}
		Zhao, Y., Lv, W., Xu, S., Wei, J., Wang, G., Dang, Q., Liu, Y., Chen, J.: DETRs beat YOLOs on real-time object detection. In: Proceedings of the IEEE/CVF Conference on Computer Vision and Pattern Recognition. pp. 8587–8596 (2024)
		
		\bibitem{Faster_R_CNN}
		Ren, S., He, K., Girshick, R., Sun, J.: Faster R-CNN: Towards real-time object detection with region proposal networks. In: Advances in neural information processing systems. pp. 91–99 (2015)
		
		\bibitem{COCO}
		Lin, T.Y., Maire, M., Belongie, S., Hays, J., Perona, P., Ramanan, D., Dollár, P., Zitnick, C.L.: Microsoft COCO: Common objects in context. In: European conference on computer vision. pp. 740–755. Springer (2014)
		
		\bibitem{DWconv}
		Chollet, F. Xception: Deep learning with depthwise separable convolutions. In Proceedings of the IEEE conference on computer vision and pattern recognition . pp. 1251-1258 (2017)
		
	\end{thebibliography}
\end{document}